\documentclass{article}
\usepackage{ijcai18}

\usepackage{times}
\usepackage{xcolor}
\usepackage{soul}
\usepackage[utf8]{inputenc}
\usepackage[small]{caption}

\usepackage{epsfig}
\usepackage{graphicx}
\usepackage{subfigure}

\usepackage{amsmath}
\usepackage{amssymb}
\usepackage{enumerate}
\graphicspath{{/}}
\usepackage{epstopdf}
\usepackage{float}

\usepackage[ruled]{algorithm2e}
\usepackage{algorithmic}

\usepackage{multirow}

\usepackage[pagebackref=true,breaklinks=true,letterpaper=true,colorlinks,bookmarks=false]{hyperref}

\title{Memory Attention Networks for Skeleton-based Action Recognition}

\begin{document}
\title{Memory Attention Networks for Skeleton-based Action Recognition}


\author{
	Chunyu Xie$^{1,a}$,
	Ce Li$^{2,a}$,
	Baochang Zhang$^{1,*}$,
	Chen Chen$^3$,
	Jungong Han$^4$,
	Changqing Zou$^5$,
	Jianzhuang Liu$^6$
	\\
	$^1$ School of Automation Science and Electrical Engineering, Beihang University, Beijing, China \\
	$^2$ Department of Computer Science and Technology, China University of Mining \& Technology, Beijing, China\\
	$^3$ Department of Electrical and Computer Engineering, University of North Carolina at Charlotte  \\
	$^4$ School of Computing \& Communications, Lancaster University, LA1 4YW, UK \\
	$^5$ University of Maryland Institute for Advanced Computer Studies, College Park, MD, 20742\\
	$^6$  Noah's Ark Lab, Huawei Technologies Co. Ltd., China\\
	yuxie@buaa.edu.cn,
	celi@cumtb.edu.cn,
	$^*$Correspondence: bczhang@buaa.edu.cn,
	chenchen870713@gmail.com,\\
	jungonghan77@gmail.com,
	cqzou@umiacs.umd.edu,
	liu.jianzhuang@huawei.com
	\thanks{Chunyu Xie and Ce Li have equal contribution to the paper.}
}
\setlength\titlebox{3in}
\maketitle
\vspace{100pt}
\begin{abstract}
	Skeleton-based action recognition task is entangled with complex spatio-temporal variations of skeleton joints, and remains challenging for Recurrent Neural Networks (RNNs). In this work, we propose a temporal-then-spatial recalibration scheme to alleviate such complex variations, resulting in an end-to-end Memory Attention Networks (MANs) which consist of a Temporal Attention Recalibration Module (TARM) and a Spatio-Temporal Convolution Module (STCM). Specifically, the TARM is deployed in a residual learning module that employs a novel attention learning network to recalibrate the temporal attention of frames in a skeleton sequence. The STCM treats the attention calibrated skeleton joint sequences as images and leverages the Convolution Neural Networks (CNNs) to further model the spatial and temporal information of skeleton data. These two modules (TARM and STCM) seamlessly form a single network architecture that can be trained in an end-to-end fashion. MANs significantly boost the performance of skeleton-based action recognition and achieve the best results on four challenging benchmark datasets: NTU RGB+D, HDM05, SYSU-3D and UT-Kinect.\footnote{The code will be made publicly available at https://github.com/memory-attention-networks.}
\end{abstract}

\section{Introduction}\label{sec:introduction}
3D skeleton-based human action recognition has recently attracted a lot of research interests due to its high-level representation and robustness to variations of viewpoints, appearances and surrounding distractions~\cite{han2016arxiv,presti20163d,Ding2016Articulated}. It is motivated by the biological observations that human beings can recognize actions from just the motion of a few joints of the human body, even without appearance information~\cite{johansson1973pp}. To describe human actions, conventional recognition approaches use relative joint coordinates to overlook the absolute movements of skeleton joints and thus gain partial view-invariant transformation. They include aligned spherical coordinates with person's direction~\cite{xia2012cvprw}, translated coordinates invariant to absolute position and orientation~\cite{jiang2015informative}, and flexible view invariant transform with principal components~\cite{raptis2011real-time}.

\begin{figure}[t]
	\vspace{-5pt}
	\begin{center}
		\includegraphics[width=0.98\linewidth]{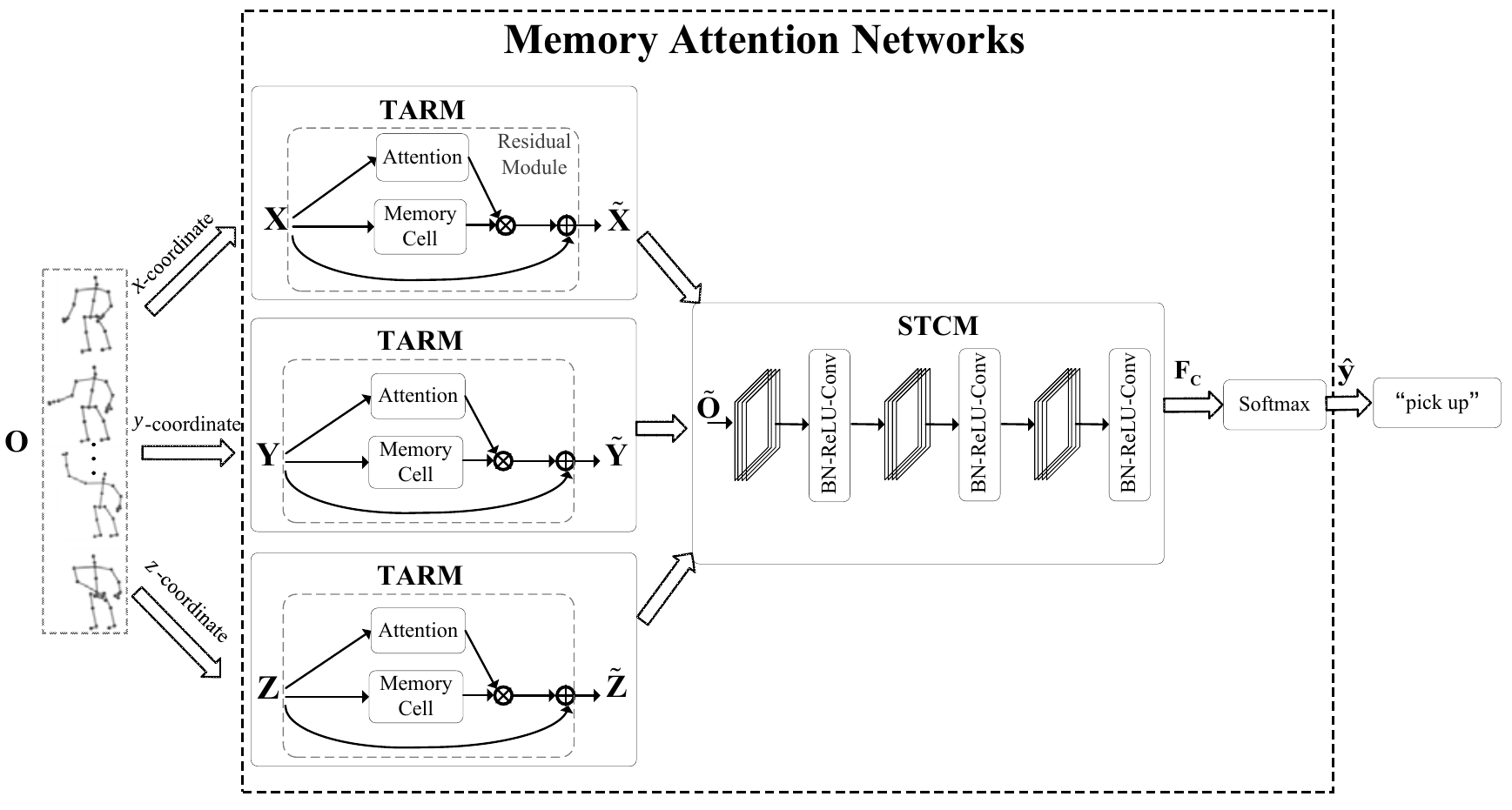}
	\end{center}
	\vspace{-5pt}
	\caption{Memory Attention Networks use the temporal-then-spatial recalibration scheme. The ({\bf TARM}) is deployed in the Residual Module (RM) to take advantage of the input features and learned attention information. The ({\bf STCM}), which treats the skeleton sequences as images and leverages the CNNs, further models the spatial and temporal information of skeleton data to cope with complex spatio-temporal variations in skeleton joints.}
	\vspace{-5pt}
	\label{fig:architecture}
\end{figure}

Skeleton sequences are time series of joint coordinate positions. To learn the temporal context of sequences, Recurrent Neural Networks (RNNs)~\cite{li2017iccv}, Long Short-Term Memory (LSTM)~\cite{zhu2016aaai}, and Gated Recurrent Unit (GRU)~\cite{cho-al-emnlp14}, have been successfully applied to skeleton based action recognition. But it still {challenging to} cope with the complex spatio-temporal variations of skeleton joints caused by a number of factors, such as action speed, jitters, and surrounding distractions. To handle these variations, attention mechanism is introduced in~\cite{liu2016eccv,liu2017cvpr,zhu2016aaai,song2017aaai} to provide a robust recognition system. {For instance, }STA-LSTM~\cite{song2017aaai} allocates different attention weights for selecting key frames and discriminative joints within one frame. Similarly, GCA-LSTM~\cite{liu2017cvpr} selects the global informative joints from a sequence.

A few works exploit CNNs to solve the skeleton based action recognition problem. In~\cite{ke2017cvpr}, skeleton joints after being projected or encoded, are used as the input channels of CNNs, which causes temporal information loss during the conversion of 3D information ($x$, $y$, $z$ joint coordinates) into 2D information (images). In~\cite{chen2017pr,lea2016eccv}, skeleton joints in each frame are transformed and expressed as color heat maps in CNNs, {where complex data preprocessing gives rise to the loss of distinct spatio-temporal information.}

\begin{table}[t]
	\caption{A brief description of notations used in the paper.}
	\vspace{-5pt}
	\renewcommand{\arraystretch}{1.3}
	\begin{center}
		\begin{tabular}{|c|l|}
			\hline
			Variable & Description  \\
			\hline\hline
			$\mathbf{O}$ & input of MANs \\
			$\mathbf{\widetilde{O}}$& output of three TARMs \\
			$\mathbf{X}$ & $x$-coordinate input of MANs \\
			$\mathbf{\widetilde{X}}$& $x$-coordinate output feature map of a TARM \\
			$\mathbf{F_M}$ & {memory information} in a TARM\\
			$\mathbf{F_A}$ & {attention weight} in a TARM\\
			$\mathbf{F_C}$ & output feature map of STCM\\
			$\hat{\mathbf{y}}$ & predicted {action label}\\ \hline
		\end{tabular}
	\end{center}
	\vspace{-5pt}
	\label{table.brief}
\end{table}
In this work, our goal is to bring these powerful tools (\emph{e.g.} RNNs, CNNs and attention learning) under the same umbrella and develop an efficient framework to investigate a new hypothesis of ``memory attention + convolution network" for skeleton based action recognition. We propose an end-to-end deep network architecture, termed as Memory Attention Networks (MANs), to perform temporal-then-spatial feature recalibration. {It can leverage the state-of-the-art CNNs to enhance the spatio-temporal features~\cite{wang2016cvprrealtime}. So far, CNNs particularly ResNets~\cite{he2016deep} or Wide-ResNets~\cite{zagoruyko2016bmvc} have been the most popular tools due to the unique residual module. Inspired by it, we design our temporal-then-spatial recalibration scheme in MANs based on the residual module as shown in Fig.~\ref{fig:architecture}. By doing so, both the original input features and the attention information can be fully exploited by subsequent CNNs in a unified framework, leading to a comprehensive and effective feature representation.}

Specifically, as shown in Fig.~\ref{fig:architecture}, each input skeleton sequence is denoted as a $T \times N \times 3$ matrix, where $T$ is the total number of frames, $N$ is the number of joints, and the 3 indicates $x$, $y$ and $z$ coordinates for each joint. For each coordinate, we have a $T \times N$ matrix. A Temporal Attention Recalibration Module (TARM) is proposed, which consists of (1) a memory cell for extracting {memory information} features by a Bidirectional Gated Recurrent Unit (BiGRU)~\cite{bahdanau2015neural} and (2) a {branch} to learn temporal attention for feature recalibration. The three temporally calibrated features $\mathbf{\widetilde{X}}$, $\mathbf{\widetilde{Y}}$ and $\mathbf{\widetilde{Z}}$ are treated as a 3-channel image and fed to a state-of-the-art CNN in the proposed Spatio-Temporal Convolution Module (STCM). In this way, the modeling ability of MANs is further enhanced by considering the spatial layout of skeleton joints. The resulting feature representations can effectively deal with the spatio-temporal variations among joints in a sequence, due to the robustness of CNNs against deformations.

\noindent {\bf {Distinctions between this work and prior art.}} (1) The state-of-the-art attention network~\cite{song2017aaai} uses two LSTMs to model spatial and temporal attentions for each skeleton frame based on the input (frames) at time steps $t$ and $t-1$. We also use RNN in TARM, but only to model the memory information of skeleton sequences. We design a new attention network to learn the attention weights and then {make use of} the learned temporal attention to recalibrate the original skeleton sequence in a residual module, which facilitates efficient learning of attention features.
(2) The existing CNNs based skeleton action recognition methods~\cite{ke2017cvpr,chen2017pr} involve complicated pre-processing. For example, \cite{ke2017cvpr} is based on clip generation (skeleton segmentation) and color images transformation; \cite{chen2017pr} performs skeleton coordinate transform and generates images with visual enhancement as the input. However, our proposed MANs directly operate on the skeleton sequences without bells and whistles, enabling an end-to-end training of network.

\noindent {\bf{Contributions.}} The contributions of this paper are threefold.
\begin{enumerate}[1.]
	\item We propose an end-to-end framework of Memory Attention Networks (MANs) to demonstrate the powerful capacity of a new ``memory attention + convolution network" scheme for modeling the complex spatio-temporal variations in skeleton joints. It is the first time that a ``RNNs + CNNs" framework has been developed for skeleton-based action recognition.
	\item {A new attention learning method is presented based on the residual module. It recalibrates temporal features to pay more attention {to} informative skeleton frames.}
	\item {MANs achieve the state-of-the-art results on four challenging datasets. We also perform extensive ablation study to show the effectiveness of each unit in MANs.}
\end{enumerate}

\section{Memory Attention Networks}\label{sec:man}
In this section, we {elaborate} the two modules: Temporal Attention Recalibration Module (TARM) and Spatio-Temporal Convolution Module (STCM) in MANs. Table~\ref{table.brief} summarizes the notations used in this paper.

\subsection{Temporal Attention Recalibration Module}\label{sec:tam}

The input skeleton data is a sequence of multi-frame 3D joint coordinates forming an action. Let $\mathbf{O}=\{\mathbf{X},\mathbf{Y},\mathbf{Z}\} \in \mathbb{R}^{T \times N \times 3}$, where $\mathbf{X} \in {{\mathbb R}^{T \times N}}$, $\mathbf{Y} \in {{\mathbb R}^{T \times N}}$, $\mathbf{Z} \in {{\mathbb R}^{T \times N}}$, denotes $N$ joints along $T$ frames with $x$, $y$, and $z$ coordinates. For ease of {explanation}, $\mathbf{X}$ is chosen as an example to describe TARM.

\begin{figure}[tbp]
	\vspace{-5pt}

	\begin{center}
		\includegraphics[width=0.98\linewidth]{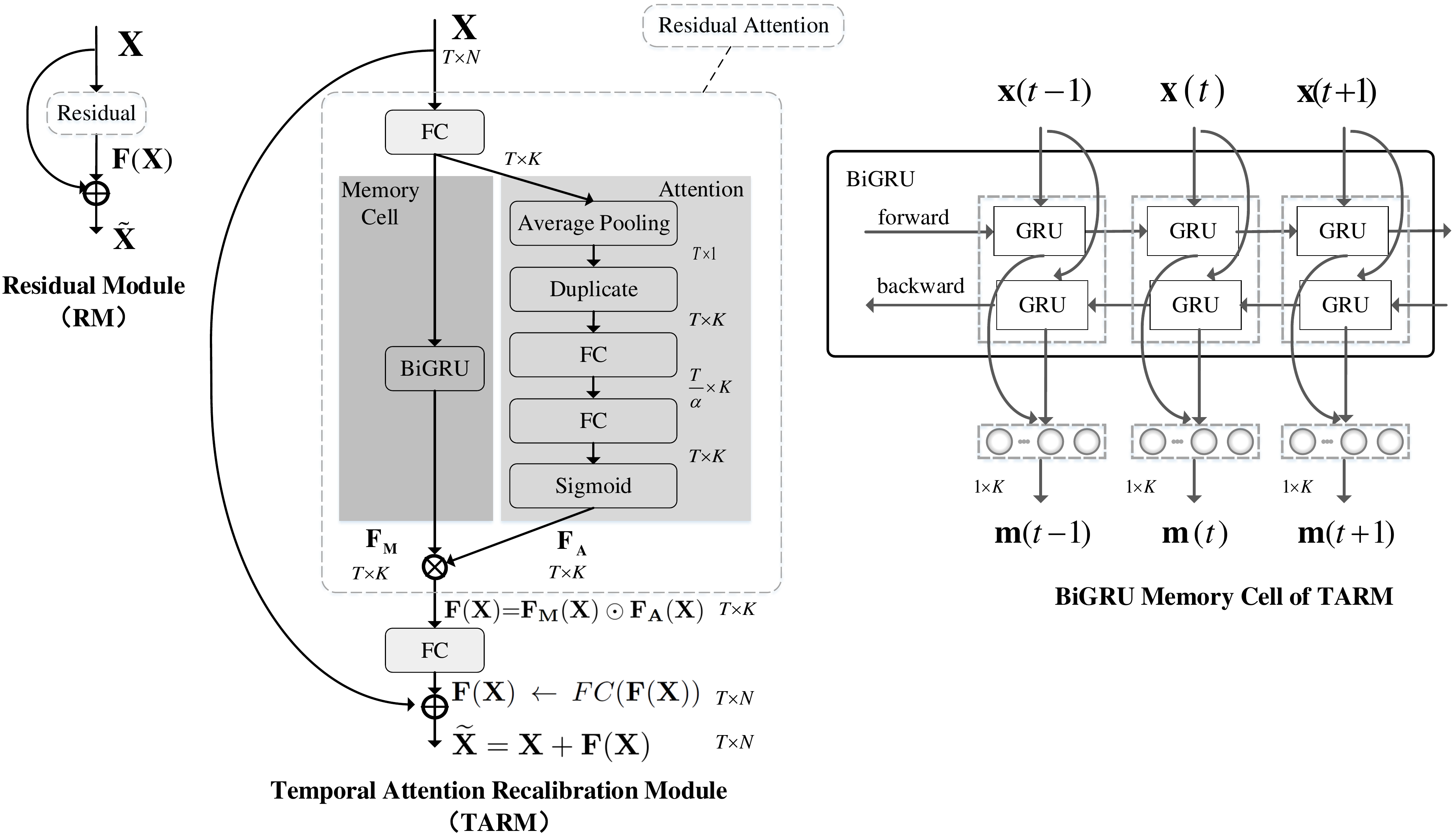}
	\end{center}
	\vspace{-5pt}
	\caption{The Residual Module ({\bf left}), Temporal Attention Recalibration Module ({\bf middle}), and BiGRU Memory Cell of TARM ({\bf right}). {TARM is designed based on the residual module as $\mathbf{\widetilde{X}} = \mathbf{X} + \mathbf{F(X)}$, which incoporates the input and recalibrated features in a unified framework. $\mathbf{F(X)}$ is the recalibrated feature, {\emph{e.g.}, the output of our residual attention module,} via  $\mathbf{F(X)}{\rm{ = }}\mathbf{{F_M}(X)} \odot \mathbf{F_A(X)}$.}}
	\label{fig:tam}
	\vspace{-5pt}
\end{figure}

As shown in Fig.~\ref{fig:tam}, given a 2D matrix $\mathbf{X}$, the learning of TARM pursuits a specific attention based on the BiGRU in memory cell to capture the temporal memory information across the input action sequence. More specifically, inspired by the original RM in ResNets, we construct TARM via identity mapping with transformation from  $\mathbf{X} \in {{\mathbb R}^{T \times N}}$  to  ${\mathbf{\widetilde{X}}} \in {{\mathbb R}^{T \times N}}$ to capture the richer temporal information, as
\begin{equation}
{	
	\mathbf{\widetilde{X}}} = \mathbf{X} + \mathbf{F(X)},
\label{eq:tam}
\end{equation}
{where $\mathbf{F(X)}$ is the recalibrated feature, {\emph{e.g.}, the output of our residual attention module shown in Fig.~\ref{fig:tam},} based on two branches: $\mathbf{F_M(X)}$ and $\mathbf{F_A(X)}$, which represent the memory information and attention weight, respectively.}
\begin{equation}
\mathbf{F(X)}{\rm{ = }}\mathbf{{F_M}(X)} \odot \mathbf{F_A(X)},
\label{eq:tamattention}
\end{equation}
where $\odot$ denotes the element-wise multiplication. 
For simplicity, $\mathbf{{F_M}(X)}$ and $\mathbf{{F_A}(X)}$ are denoted as $\mathbf{{F_M}}\in \mathbb{R}^{T \times K}$ and $\mathbf{{F_A}} \in \mathbb{R}^{T \times K}$, respectively. $\mathbf{{F_A}}$ is the weight of $\mathbf{{F_M}}$ to recalibrate {temporal information. }
Obviously, for an action sequence, the importance of representative information in each frame is different, {and only a few key frames containing important discriminative information deserves to be emphasized for action representation.}


\vspace{-1em}\paragraph{{Calculating $\mathbf{{F_M}}$}.}
We {implement} the memory cell via BiGRU. $\mathbf{X} \in \mathbb{R}^{T \times N}$ is resized and updated by the output of a FC layer as $\mathbf{X} \leftarrow FC(\mathbf{X}) \in \mathbb{R}^{T \times K}$, as shown at the top of Fig.~\ref{fig:tam}. In a slight abuse of notation, we still denote the resized output as $\mathbf{X}$. $\mathbf{{F_M}} \in \mathbb{R}^{T \times K}$ {is the memory information made up of }two directional combined hidden states in BiGRU, where $K$ denotes the number of neuron units in BiGRU.

For simplicity, we still denote $\mathbf{{X}}\in \mathbb{R}^{T \times K}$ as
\begin{equation}
\scalebox{0.85}{$\mathbf{X}=\left[ {\begin{array}{*{20}{c}}
		{\mathbf{x}(1)}\\
		\vdots \\
		{\mathbf{x}(t)}\\
		\vdots \\
		{\mathbf{x}(T)}
		\end{array}} \right] = \left[ {\begin{array}{*{20}{c}}
		{{x_1}(1)}& \cdots &{{x_k}(1)}& \cdots &{{x_K}(1)}\\
		\vdots & \ddots &{}&{}& \vdots \\
		{{x_1}(t)}&{}&{{x_k}(t)}&{}&{{x_K}(t)}\\
		\vdots &{}&{}& \ddots & \vdots \\
		{{x_1}(T)}& \cdots &{{x_k}(T)}& \cdots &{{x_K}(T)}
		\end{array}} \right]$},
\label{eq:xarrary}
\end{equation}
where ${\mathbf{x}(t)}$ is a row vector of $\mathbf{X}$ to represent the sequence at $t^{th}$ frame as $\left[{{x_1}(t)}, \cdots  ,{{x_K}(t)}\right]$, and $t \in \left({1,...,T} \right)$.

As illustrated in Fig.~\ref{fig:tam}, $\mathop {\mathbf{GRU}}\limits^ {\scalebox{3}[0.5]{$\rightarrow$}}  {\rm{(}}\mathbf{x}(t){\rm{)}}$ and $\mathop {\mathbf{GRU}}\limits^ {\scalebox{3}[0.5]{$\leftarrow$}}  {\rm{(}}\mathbf{x}(t){\rm{)}}$ are the output {hidden states} of the forward GRU and backward GRU, respectively.
{Combining these bidirectional hidden states, the informative vector $\mathbf{m}(t)\in \mathbb{R}^{1 \times K}$ at the $t^{th}$ frame is denoted as
	\begin{equation}
	\mathbf{m}(t) = \mathop {\mathbf{GRU}}\limits^ {\scalebox{3}[0.5]{$\rightarrow$}}  {\rm{(}}\mathbf{x}(t){\rm{)}} + \mathop {\mathbf{GRU}}\limits^ {\scalebox{3}[0.5]{$\leftarrow$}}  {\rm{(}}\mathbf{x}(t){\rm{)}}.
	\label{eq:bigru}
	\end{equation}
	Finally, all the outputs are concatenated across $T$ frames, and the memory information of skeleton joints is represented as
}
\begin{equation}
{\mathbf{F_M}}={\left[ {\mathbf{m}(1), \cdots ,\mathbf{m}(T}) \right]^T\in \mathbb{R}^{T \times K}},
\label{eq:memory}
\end{equation}
where $\mathbf{F_M}$ summarizes the memory information in BiGRU for the skeleton joints across the sequence.

\vspace{-1em}\paragraph{{Calculating $\mathbf{{F_A}}$}.}
{To recalibrate the memory information  $\mathbf{F_M}$, the attention weight $\mathbf{{F_A}}$ is exploited as shown in Eq.~(\ref{eq:tamattention}). In Fig.~\ref{fig:tam}, following $\mathbf{X} \in {\mathbb{R}^{T \times K}}$, our recalibration scheme can capture  global frame-wise dependence across $T$ frames. }{We first aggregate each row vector of $\mathbf{X}$ in Eq.~(\ref{eq:xarrary}) by the average pooling operation to produce a $T \times 1$ vector as
	\begin{equation}
	{{\mathbf{X}}_{{p}}} = {\left[ {\frac{1}{K}\sum\limits_{k = 1}^K {{x_k}\left( 1 \right)} , \cdot  \cdot  \cdot ,\frac{1}{K}\sum\limits_{k = 1}^K {{x_k}\left( T \right)} } \right]^T} \in {\mathbb{R}^{T \times 1}}.
	\label{eq:averagepooling}
	\end{equation}
	We then duplicate it with $K$ copies as
	\begin{equation}
	{\mathbf{X} \leftarrow {\mathbf{F}_d}\left( {\mathbf{X}_p,K} \right)\in {\mathbb{R}^{T \times K}}},
	\label{eq:x2}
	\end{equation}
	where ${\mathbf{F}_d}\left( {\mathbf{X}_p,K} \right)=\underbrace {\left[ {{{\mathbf{X}}_{{p}}}, \cdots ,{{\mathbf{X}}_{{p}}}} \right]}_K$.
}

In the right branch of TARM shown in Fig.~\ref{fig:tam}, the {attention} mechanism is represented by a bottleneck with the two FC layers providing the non-linear interaction between frames. We introduce a dimensionality-reduction layer with parameters $\mathbf{W}_1$ and a ratio factor $\alpha$ (empirically set to be 16 in Section~\ref{sec:implementation}), followed by a ReLU function. And then, we introduce a dimensionality-increasing layer with parameters $\mathbf{W}_2$ and a sigmoid activation function. {The dimensionality-reduction and dimensionality-increasing processing can be considered as the denoising  and excitation operations respectively, and thus enhance the feature discriminability.} Finally, the output of the attention branch $\mathbf{F_A}$ is calculated as
\begin{equation}
\mathbf{F_A} = \sigma \left( {{\mathbf{W}_2}\;\theta\left({{\mathbf{W}_1}\mathbf{X}}\right) } \right) \in {\mathbb{R}^{T \times K}},
\label{eq:attention}
\end{equation}
where ${\mathbf{W}_1} \in {{\mathbb R}^{\frac{T}{\alpha } \times T}}$ and ${\mathbf{W}_2} \in {{\mathbb R}^{T \times \frac{T}{\alpha }}}$ . To simplify the notation, the bias terms in Eq.~(\ref{eq:attention}) are omitted. $\theta \left(  \cdot  \right)$ refers to the ReLU function, and $\sigma \left(  \cdot  \right)$ denotes the sigmoid function. Finally, $\mathbf{F}(\mathbf{X})$ is obtained by the element-wise multiplication of $\mathbf{F_M}$ and $\mathbf{F_A}$. 

Furthermore, to calculate the output feature map of TARM, $\mathbf{\widetilde{X}}$,
$\mathbf{F}(\mathbf{X}) \in \mathbb{R}^{T \times K}$ is resized by a FC layer as $\mathbf{F}(\mathbf{X}) \leftarrow FC(\mathbf{F}(\mathbf{X})) \in \mathbb{R}^{T \times N}$,  {shown at the bottom of Fig.~\ref{eq:tamattention}}. {As a result,} the final  $\mathbf{F}(\mathbf{X})$ describes the temporal information of the entire skeleton sequence. Similar to RM in ResNets, $\mathbf{F_M}$ and $\mathbf{F_A}$ in TARM can be jointly learned during training. In a similar way, we can obtain $\mathbf{\widetilde{Y}}$ and $\mathbf{\widetilde{Z}}$ based on $\mathbf{{Y}}$ and $\mathbf{{Z}}$ in their corresponding TARM, and $\mathbf{\widetilde{O}}=\{\mathbf{\widetilde{X}},\mathbf{\widetilde{Y}},\mathbf{\widetilde{Z}}\}\in \mathbb{R}^{T \times N \times 3}$.

\subsection{Spatio-Temporal Convolution Module}\label{sec:scm}
Conventional attention methods in skeleton action recognition are limited by the modeling capacity of RNNs~\cite{liu2016eccv,liu2017cvpr}.
STCM is introduced based on CNNs to extract the enhanced spatio-temporal features from the output ($\mathbf{\widetilde{X}}$, $\mathbf{\widetilde{Y}}$ and $\mathbf{\widetilde{Z}}$) of the TARM. By leveraging the robustness to deformation of CNNs, STCM further extracts high-level feature representations to better cope with spatio-temporal variations of skeleton joints.

\begin{table*}[t]
	\vspace{4pt}
\small
	\caption{The architectures of various MANs-$n$ (\emph{i.e.}, $n$ = 9, 33, 61).}
	\renewcommand{\arraystretch}{1.4}
	\vspace{-6pt}
	\centering
	\begin{tabular}{c|c|c|c|c}
		\hline\hline
		Module&Output Size & \multicolumn{1}{c}{MANs-9} & \multicolumn{1}{c}{MANs-33} & \multicolumn{1}{c}{MANs-61} \\ \hline
		TARM&$50 \times 50$ & \multicolumn{3}{c}{$64 \times 2$}\\
		\hline
		\multirow{6}[0]{*}{STCM}&$25 \times 25$ & \multicolumn{3}{c}{$5 \times 5$, 64, stride 2 }\\
		\cline{2-5}
		& $25 \times 25$&
		$\left[ {\begin{array}{*{20}{c}}
			3 \times 3, 64\\
			3 \times 3, 64
			\end{array}} \right] \times 2 $&$\left[ {\begin{array}{*{20}{c}}
			3 \times 3, 64\\
			3 \times 3, 64
			\end{array}} \right] \times 8 $&$\left[ {\begin{array}{*{20}{c}}
			3 \times 3, 64\\
			3 \times 3,64
			\end{array}} \right] \times 15 $\\
		\cline{2-5}
		&$13 \times 13$&
		$\left[ {\begin{array}{*{20}{c}}
			3 \times 3, 128\\
			3 \times 3, 128
			\end{array}} \right] \times 2 $&$\left[ {\begin{array}{*{20}{c}}
			3 \times 3, 128\\
			3 \times 3, 128
			\end{array}} \right] \times 8 $&$\left[ {\begin{array}{*{20}{c}}
			3 \times 3, 128\\
			3 \times 3,128
			\end{array}} \right] \times 15 $\\
		\cline{2-5}
		&$1 \times 1$&\multicolumn{3}{c}{Average Pooling, FC, Softmax}\\
		\hline\hline
	\end{tabular}%
	\vspace{-10pt}
	\label{tab:architecture}%
\end{table*}%

In principal, any CNNs can be used in STCM, \emph{e.g.}, DenseNets~\cite{huang2017densely} and ResNets~\cite{he2016deep}. ${\mathbf{\widetilde{O}}} \in \mathbb{R}^{T \times N \times 3}$ denotes the output of TARMs and also the input to STCM, and $\mathbf{F_C}$ denotes the output of STCM for the softmax classifier. For example, in Fig.~\ref{fig:architecture}, the BN-ReLU-Conv blocks in STCM are used to interpret the high-level spatial structures of skeleton joints as
\begin{equation}
\scalebox{0.9}{$\mathbf{F_C}={\rm {Conv}}\left({\rm{ReLU}}\left({\rm{BN}}\left(...{\rm{Conv}}\left({\rm{ReLU}}\left({\rm {BN}}\left({\mathbf{\widetilde{O}}}\right)\right)\right)\right)\right)\right)$}.
\label{eq:convolution}
\end{equation}
Afterwards, $\mathbf{F_C}$ is fed to a softmax classifier to predict the class label as
\begin{equation}
\hat{\mathbf{y}} = {\rm{softmax}}\left( {{\mathbf{W}_C},\mathbf{F_C}} \right),
\label{eq:softmax}
\end{equation}
where $\mathbf{W}_C$ and $\hat{\mathbf{y}}$ denote the weights in the softmax layer and the predicted action label, respectively. The cross-entropy loss function~\cite{goodfellow2016dl} is adopted to measure the difference between the true class label ${\mathbf{y}}$ and the prediction result $\hat{\mathbf{y}}$. 

\section{Experiments}\label{sec:experiments}

The proposed MANs are evaluated on four public skeleton action datasets: NTU RGB+D~\cite{sharhroudy2016cvpr}, HDM05~\cite{muller2005hdm05}, SYSU-3D~\cite{hu2015cvpr} and UT-Kinect~\cite{xia2012cvprw}.
%
%
%
%

\subsection{Datasets and Implementation}\label{sec:implementation}
{\bf NTU RGB+D dataset.} The NTU dataset~\cite{sharhroudy2016cvpr} is the largest skeleton-based action recognition dataset, with more than 56000 sequences and 4 million frames. There are 60 classes of actions performed by 40 subjects.
In total, there are 80 views for this dataset, and each skeleton has 25 joints. Due to the large viewpoint, intra-class and sequence length variations, the dataset is very challenging. For fair comparisons, we follow the same cross-subject and cross-view evaluation protocols in~\cite{sharhroudy2016cvpr}.

\noindent{\bf HDM05 dataset.} The HDM05 dataset~\cite{muller2005hdm05} contains 2,337 skeleton sequences performed by 5 actors (613,377 frames). We use the same experiment setting (65 classes, 10-fold cross validation) in~\cite{zhu2016aaai}.

\noindent{\bf SYSU-3D dataset.} The SYSU-3D dataset~\cite{hu2015cvpr} collected with the Microsoft Kinect contains 12 actions performed by 40 subjects.
The dataset has 480 skeleton sequences and is very challenging as the motion patterns are quite similar among different action classes. Moreover, there are a lot of viewpoint variations. We evaluate the performance of our method using the standard 30-fold cross-validation protocol~\cite{hu2015cvpr}, in which half of the subjects are used for training and the rest for testing.

\noindent{\bf UT-Kinect dataset.} The UT-Kinect dataset~\cite{xia2012cvprw} is collected using a single stationary Kinect. The skeleton sequences in this dataset are very noisy. 10 action classes are performed by 10 subjects, and each action is performed by the same subject twice. We follow the standard Leave-One-Out-Cross-Validation (LOOCV) protocol in~\cite{xia2012cvprw}.

\noindent{\bf Implementation details.}\label{sec:setting} For all the datasets, the matrices ($\mathbf{\widetilde{X}}$, $\mathbf{\widetilde{Y}}$ and $\mathbf{\widetilde{Z}}$) are generated with all the frames of a skeleton sequence. {We use two different scales for the three input matrices of each sequence, \emph{i.e.},  $224 \times 224$ and $50 \times 50$, respectively. For the large scale, the number of  hidden units of BiGRU in TARM is set to $2 \times 128$ ($K=128$), where $2$ indicates bidirectional GRU, $128$ is the number of neurons. DenseNet-161~\cite{huang2017densely}  and  ResNet-18~\cite{he2016deep}  are used in STCM, leading to MANs (DenseNet-161) or MANs (ResNet-18). For the small scale, we set the hidden units of BiGRU to $2 \times 64$, and stack multiple BN-ReLU-Conv blocks as STCM, resulting in MANs-$n$ (\emph{e.g.} $n=9$) where $n$ is the number of BN-ReLU-Conv blocks. The architectures of various MANs are illustrated in Table~\ref{tab:architecture}.}

The number of the units for the last FC layer (\emph{i.e.}, the output layer) is the same as the number of the action classes in each dataset. MANs are trained using the stochastic gradient descent algorithm, and the learning rate, decay, and momentum, are respectively set to 0.1, 0, and 0.9. The mini-batches of samples on NTU RGB+D, HDM05, SYSU-3D, and UT-Kinect are constructed by randomly sampling 40, 20, 8, and 8 samples from the training sets, respectively. The training stops after 100 epochs except for NTU RGB+D after 50 epochs. For a fair comparison, the performance of MANs on each dataset is compared with existing methods using the same evaluation protocol. All experiments are performed based on Keras\footnote{http://keras.io} with Tensorflow backend using two NVIDIA Titan X Pascal GPUs.
\subsection{Experiment Analysis}\label{sec:analysis}

\noindent {\bf Parameter analysis.} To investigate the performance of MANs using different values of the ratio factor $\alpha$ in Eq.~(\ref{eq:attention}), the comparison of accuracy on the same trial of HDM05 dataset is conducted by MANs (ResNet-18) in Table~\ref{tab:alphak}. From the second column to the fifth column of Table~\ref{tab:alphak}, the results show that MANs (ResNet-18) consistently keep high training efficiency using $\alpha=4,8,16,32$. MANs (ResNet-18) with the ratio factor $\alpha=16$ achieve the best accuracy of 99.23\% on the HDM05 dataset.  The parameter tuning experiment reveals that $\alpha=16$ is a proper ratio factor for generating {the attention weight.} In all the following experiments, we evaluate the performance of MANs by setting $\alpha$ to 16.

\begin{table}[tbp]
	\caption{Recognition accuracies of MANs (ResNet-18) on the HDM05 dataset using  $\alpha$ = 4, 8, 16, 32.}
	\renewcommand{\arraystretch}{1.2}
\small
	\centering
	\vspace{-6pt}
	\begin{tabular}{ccccc}
		\hline\hline
		{$\alpha $} & 4 & 8 & 16 &32 \\
		\hline
		Accuracy(\%) & 98.09 & 98.38 & {\bf 99.23} & 98.64 \\
		\hline\hline
	\end{tabular}%
	\label{tab:alphak}%
	\vspace{-3pt}
\end{table}

\begin{table}[tbp]
	\caption{Performance of STCM, MANs (no attention), MANs (other temporal attention) and MANs on the NTU RGB+D dataset.}
	\vspace{-6pt}
	\centering
	\resizebox{!}{0.12\textwidth}{
		\renewcommand{\arraystretch}{1.3}
		\begin{tabular}{l c c}
			\hline\hline
			Method & CS. & CV.\\
			\hline
			STCM-9 (CNNs)  & 81.31   &89.78      \\
			MANs-9 (no attention) &81.41    &89.84       \\
			MANs-9 (other temporal attention)  &81.94   &90.12   \\
			MANs-9 & \bf{83.01}   &  \bf{90.66}    \\
			\hline
			STCM (DenseNet-161) & 81.56   &  90.24   \\
			MANs (DenseNet-161, no attention) &81.96 & 92.15 \\
			MANs (DenseNet-161, other temporal attention)&81.60 & 92.18  \\
			MANs (DenseNet-161)  & \bf{82.67}   &  \bf{93.22}     \\
			\hline
			\hline
		\end{tabular}%
	}
	\vspace{-10pt}
	\label{tab:ablation}%
\end{table}

\begin{table*}[tbp]
	\centering
	\small
	\renewcommand\arraystretch{1.25}
	\caption{Comparison of the results of different units in MANs on four datasets.}
	\vspace*{-8pt}
	\begin{tabular}{l c c c c c c}
		\hline\hline
		\multirow{2}[4]{*}{Method}&\multirow{2}[4]{*}{\#param} & \multicolumn{2}{c}{{NTU RGB+D}} & \multirow{2}[4]{*}{{HDM05}} & \multirow{2}[4]{*}{{SYSU-3D}} &  \multirow{2}[4]{*}{{UT-Kinect}} \\
		\cline{3-4}    &      & Cross Subject & Cross View &       &       &         \\
		\hline		
		Hierarchical RNNs~\cite{du2015hierarchical}&- & 59.10 & 64.00 & 96.92 &-  &-\\
		Dynamic skeletons~\cite{hu2015cvpr}&- & 60.23 & 65.22 &- &75.50 &- \\
		Deep LSTM~\cite{zhu2016aaai}&0.6M & - & - & 96.80 &- &- \\
		ST-LSTM~\cite{liu2016eccv}&- & 61.70  & 75.50 &- &76.50  &97.00\\
		ST-LSTM + TG~\cite{liu2016eccv}&- & 69.20 & 77.70 &- &76.80  &97.50 \\
		Two-stream RNNs~\cite{wang2017cvprtwostream}&  & 71.30  & 79.50 &- &-  &- \\
		STA-LSTM~\cite{song2017aaai}&0.5M & 73.40  & 81.20 &- &-  &-\\
		Adaptive RNN-T~\cite{li2017iccv}&- & 74.60 & 83.20 &- &-  &-\\
		GCA-LSTM~\cite{liu2017cvpr}&- & 76.10  & 84.00 &- &78.60  &99.00\\
		Clips+CNN+MTLN~\cite{ke2017cvpr}&62M & 79.57 & 84.83 &- &-  &-\\
		VA-LSTM~\cite{zhang2017iccv}&- & 79.40  & 87.60 &- &77.50  &-\\
		\hline
		MANs-9 &0.8M & \bf{83.01}   &  90.66  &  98.46    &  87.04        & \bf{100.0}  \\
		MANs-33 &3.1M &82.40    &90.94    & \bf{98.85}     &   86.81     &100.0   \\
		MANs-61 &5.7M &82.42    &\bf{90.97}    &  98.76   &    \bf{87.63}   & 99.50  \\
		\hline
		MANs (ResNet-18)&12.0M  & 79.74   &  91.55   &  \bf{99.04}    &  \bf{86.93}       & \bf{100.0}  \\
		MANs (DenseNet-161)&27.6M  & \bf{82.67}   &  \bf{93.22}   &  97.69    &  78.86      &  99.00   \\
		\hline\hline
	\end{tabular}%
	\label{tab:results}%
	\vspace*{-8pt}
\end{table*}%

\noindent \textbf{Ablation study.} We conduct extensive ablation study of different units in MANs with the following settings: (A) STCM\footnote{Here STCM takes [$\mathbf{X}, \mathbf{X}, \mathbf{Z}$] as image input to perform classification via CNNs.} (\emph{i.e.} applying CNN on the original skeleton images); (B) MANs (no attention)–MANs without attention in TARM (\emph{i.e.} no $\mathbf{F_A}$); (C) MANs (other temporal attention)–MANs use the temporal attention scheme in~\cite{song2017aaai} for the TARM. Table~\ref{tab:ablation} shows the results of different architectures on the NTU RGB+D dataset. {Note that the STCM of MANs uses DenseNet-161 for the input of $224 \times 224$, and uses stacked BN-ReLU-Conv blocks with 9 layers for the input of $50 \times 50$ in Table~\ref{tab:ablation}, respectively.}
Comparing with our full MANs model, we have these findings. (1) Setting A yileds much lower performance, indicating the importance of temporal information modeling for skeleton. (2) Setting B reveals the effectiveness of the proposed attention mechanism (\emph{i.e.} learning $\mathbf{F_A}$). (3) Setting C substitutes the attention scheme in~\cite{song2017aaai} for our residual attention module in MANs. The recognition accuracy is lower than that of our MANs, which again validates the superiority of our residual attention learning approach.

{
\noindent \textbf{Learning convergence.} We plot the training error and testing error curves of the four networks on a same trail of NTU RGB+D dataset in Fig.~\ref{fig:convergence}, including STCM-9, MANs-9 (no attention), MANs-9 (other temporal attention), and MANs-9. We can observe that MANs-9 in solid line converges at epoch \#26 for training and obtains the best error rate of 16.99\% for testing. For training error, STCM-9 stops decreasing at epoch \#30, MANs-9 (no attention) converges at epoch \#30, and MANs-9 (other temporal attention) converges at epoch \#40. These curves show that MANs obviously converge faster and gain better performance than others. For example, MANs converge quickly and improve the performance over STCM, which proves that the ``memory attention + convolution network" scheme in MANs can be used to improve the modeling ability of CNNs. MANs-9 converges much faster than MANs-9 (other temporal attention) (epoch \#26 {v.s.} \#40), due to the novel residual attention module which takes the input and attention information into account in the same framework. More specifically, the residual attention module not only uses the temporal attention recalibrated information, but also delivers the spatial structure information of the original input by the identity shortcut.

%

\begin{figure}[tbp]
	\vspace{-5pt}
	\centering
	\subfigure[Training error v.s. epoch]{
		\centering
		\includegraphics[width=0.48\linewidth]{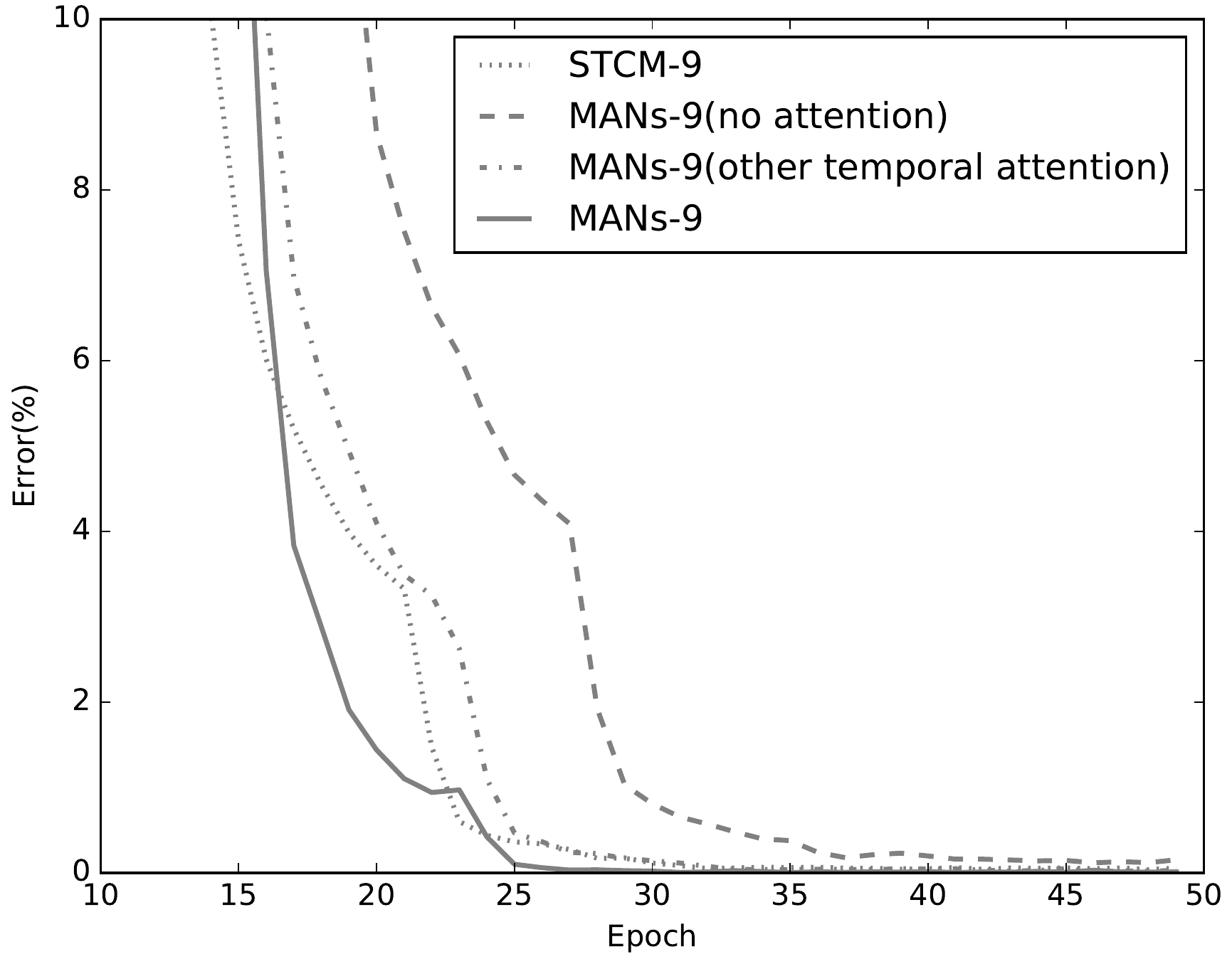}}
	\subfigure[Testing error v.s. epoch]{
		\centering
		\includegraphics[width=0.48\linewidth]{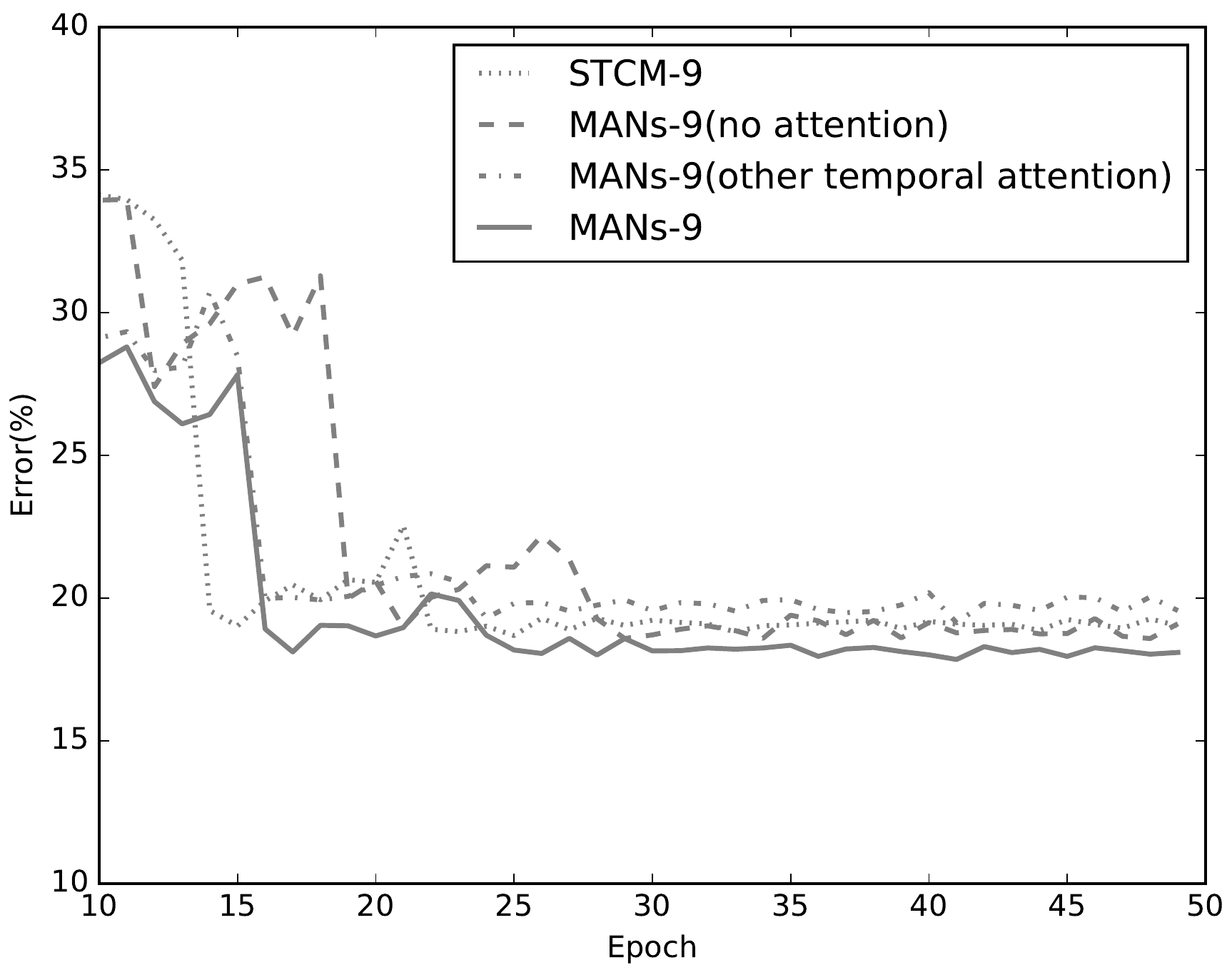}}
	\vspace{-7pt}
	\caption{Training and testing error curves of STCM-9, MANs-9 (no attention), MANs-9 (other temporal attention) and MANs-9 on the NTU RGB+D dataset (cross-subject setting). }
	\label{fig:convergence}
	\vspace{-12.5pt}
\end{figure}


{
\noindent \textbf{Various CNNs in STCM.} The number of stacked layers of CNNs in STCM, \emph{i.e.}, multiple BN-ReLU-Conv blocks used in MANs-9, MANs-33 and MANs-61, is evaluated in terms of the recognition accuracy on all the datasets (Table~\ref{tab:results}). We also include MANs (ResNet-18), MANs (DenseNet-161) in the same table. The deeper CNNs have more learnable parameters than the shallower ones. We note that MANs-9 \textbf{with a similar parameter amount} as Deep LSTM~\cite{zhu2016aaai} has much better performance than the state-of-the-arts. This reveals that our method is more effective if the network complexity should be considered. Interestingly, MANs-9 mostly achieve better performance than other deeper models, which is probably due to its compactness. However, the very deep MANs (DenseNet-161) obtain the best result on the challenging NTU RGB+D dataset in cross-view setting. It is worth noting that because of the ``memory attention + convolution network" scheme of MANs, it is quite flexible to deploy different parameter amount in MANs by adjusting the number of CNN layers in STCM to balance between performance and network complexity.

\subsection{Results and Comparisons}\label{sec:results}
We show the performance comparison of various MANs architectures with other state-of-the-art approaches in Table~\ref{tab:results} for the four datasets, respectively. {Note that all our MANs under different parameter amounts are able to achieve better performance than the state-of-the-art RNNs-based approaches (\emph{e.g.} VA-LSTM~\cite{zhang2017iccv}, GCA-LSTM~\cite{liu2017cvpr}) and the state-of-the-art CNNs-based approaches (\emph{e.g.} Clips+CNN+MTLN~\cite{ke2017cvpr}), demonstrating the superiority of our ``memory attention + convolution network" architecture.}

We analyze the best results of various MANs for the NTU RGB+D dataset. MANs perform significantly better than others in both the cross-subject and cross-view protocols. The accuracies of MANs-9 are 83.01\% for the cross-subject protocol and 90.66\% for the cross-view protocol. MANs (DenseNet-161) achieve {82.67\%} in the cross-subject test and {93.22\%} in the cross-view test. Comparing to other methods, MANs-9 increase the accuracy by {3.44\%} for cross-subject evaluation, and MANs (DenseNet-161) lead to a significant {5.62\%} improvement on this largest dataset in cross-view evaluation, which demonstrate that MANs can learn more discriminative spatio-temporal features to alleviate the spatial and temporal variations in skeleton joints.

For the HDM05 dataset, our MANs-33 {achieves} better result than the state-of-the-art multi-layer RNNs-based models. Our MANs (ResNet-18) achieve even better result up to 99.04\%. The improved results of MANs-33 and MANs (ResNet-18) suggest that CNNs in STCM not only enhance the temporal attention information in TARM, but also exhibit better motion modeling ability than deep RNNs-based model by the flexible architecture. For the SYSU-3D dataset, five accuracies of our MANs are higher than all the RNNs-based methods~\cite{hu2015cvpr,liu2016eccv,liu2017cvpr,zhang2017iccv}; especially MANs-61 outperform the previous best approach GCA-LSTM by {\bf 9.03\%}. It validates the superiority of MANs in skeleton based action recognition, and that the temporal-then-spatial recalibration scheme is effective for this task which suffers from lots of variations. For the UT-Kinect datasest, MANs-9 and MANs (ResNet-18) achieve a 100\% accuracy, with 1.0\% improvement in comparison with the state-of-the-art GCA-LSTM. This experiment shows that compared with the existing RNNs-based methods, the deployed residual module with temporal attention and the temporal-then-spatial scheme can effectively improve the modeling ability of RNNs.

}
\section{Conclusion}\label{sec:conclusion}
{In this paper, we propose an end-to-end framework, termed Memory Attention Networks (MANs), to enhance the spatio-temporal features for skeleton-based action recognition. In MANs, TARM is designed  to recalibrate the temporal attention to skeleton frames in action sequences, and STCM further models the spatial structure and temporal dependence of the skeleton sequence by leveraging the powerful CNNs. Through the unified framework, MANs significantly boost the performance for skeleton-based action recognition. The extensive experiments validate the superiority of MANs, which consistently perform the best on four benchmark datasets and contribute new state-of-the-art results. 
\section*{Acknowledgement}
The work was supported by the Natural Science Foundation of China under Contract 61672079, 61473086, and 61601466, the Open Projects Program of National Laboratory of Pattern Recognition, and Shenzhen Peacock Plan KQTD2016112515134654. Baochang Zhang is also with Shenzhen Academy of Aerospace Technology, Shenzhen, China. 
	

\bibliographystyle{named}
\bibliography{egbib}

\begin{thebibliography}{}

\bibitem[\protect\citeauthoryear{Bahdanau \bgroup \em et al.\egroup
  }{2015}]{bahdanau2015neural}
Dzmitry Bahdanau, Kyunghyun Cho, and Yoshua Bengio.
\newblock Neural machine translation by jointly learning to align and
  translate.
\newblock In {\em ICLR}, 2015.

\bibitem[\protect\citeauthoryear{Cho \bgroup \em et al.\egroup
  }{2014}]{cho-al-emnlp14}
Kyunghyun Cho, Bart van Merri{\"{e}}nboer, {\c C}ağlar G{\"{u}}l{\c c}ehre,
  Dzmitry Bahdanau, Fethi Bougares, Holger Schwenk, and Yoshua Bengio.
\newblock Learning phrase representations using {RNN} encoder--decoder for
  statistical machine translation.
\newblock In {\em Empirical Methods in Natural Language Processing}, pages
  1724--1734, 2014.

\bibitem[\protect\citeauthoryear{Ding and Fan}{2016}]{Ding2016Articulated}
Meng Ding and Guoliang Fan.
\newblock Articulated and generalized gaussian kernel correlation for human
  pose estimation.
\newblock {\em IEEE Transactions on Image Processing A Publication of the IEEE
  Signal Processing Society}, 25(2):776, 2016.

\bibitem[\protect\citeauthoryear{Du \bgroup \em et al.\egroup
  }{2015}]{du2015hierarchical}
Yong Du, Wei Wang, and Liang Wang.
\newblock Hierarchical recurrent neural network for skeleton based action
  recognition.
\newblock In {\em CVPR}, pages 1110--1118, 2015.

\bibitem[\protect\citeauthoryear{Goodfellow \bgroup \em et al.\egroup
  }{2016}]{goodfellow2016dl}
Ian Goodfellow, Yoshua Bengio, and Aaron Courville.
\newblock {\em Deep Learning}.
\newblock MIT Press, 2016.
\newblock \url{http://www.deeplearningbook.org}.

\bibitem[\protect\citeauthoryear{Han \bgroup \em et al.\egroup
  }{2017}]{han2016arxiv}
Fei Han, Brain Reily, William Hoff, and Hao Zhang.
\newblock Space-time rep-resentation of people based on 3{D} skeletal data: A
  review.
\newblock {\em Computer Vision and Image Understanding}, 158:85--105, 2017.

\bibitem[\protect\citeauthoryear{He \bgroup \em et al.\egroup
  }{2016}]{he2016deep}
Kaiming He, Xiangyu Zhang, Shaoqing Ren, and Jian Sun.
\newblock Deep residual learning for image recognition.
\newblock In {\em CVPR}, pages 770--778, 2016.

\bibitem[\protect\citeauthoryear{Hu \bgroup \em et al.\egroup
  }{2015}]{hu2015cvpr}
Jianfang Hu, Weishi Zheng, Jianhuang Lai, and Jianguo Zhang.
\newblock Jointly learning heterogeneous features for {RGB-D} activity
  recognition.
\newblock In {\em CVPR}, pages 5344--5352, 2015.

\bibitem[\protect\citeauthoryear{Huang \bgroup \em et al.\egroup
  }{2017}]{huang2017densely}
Gao Huang, Zhuang Liu, Laurens van~der Maaten, and Kilian~Q Weinberger.
\newblock Densely connected convolutional networks.
\newblock In {\em CVPR}, 2017.

\bibitem[\protect\citeauthoryear{Jiang \bgroup \em et al.\egroup
  }{2015}]{jiang2015informative}
Min Jiang, Jun Kong, George Bebis, and Hongtao Huo.
\newblock Informative joints based human action recognition using skeleton
  contexts.
\newblock {\em Signal Processing: Image Communication}, 33:29--40, 2015.

\bibitem[\protect\citeauthoryear{Johansson}{1973}]{johansson1973pp}
Gunnar Johansson.
\newblock Visual perception of biological motion and a model for its analysis.
\newblock {\em Perception and Psychophysic}, 14(2):201--211, 1973.

\bibitem[\protect\citeauthoryear{Ke \bgroup \em et al.\egroup
  }{2017}]{ke2017cvpr}
Qiuhong Ke, Mohammed Bennamoun, Senjian An, Ferdous Sohel, and Farid Boussaid.
\newblock A new representation of skeleton sequences for 3{D} action
  recognition.
\newblock In {\em CVPR}, 2017.

\bibitem[\protect\citeauthoryear{Lea \bgroup \em et al.\egroup
  }{2016}]{lea2016eccv}
Colin Lea, Reiter~Austin Vidal, Ren\'e, and Gregory~D. Hager.
\newblock Temporal convolutional networks: A unified approach to action
  segmentation.
\newblock {\em ECCV}, pages 47--54, 2016.

\bibitem[\protect\citeauthoryear{Li \bgroup \em et al.\egroup
  }{2017}]{li2017iccv}
Wenbo Li, Longyin Wen, Ming-Ching Chang, Ser~Nam Lim, and Siwei Lyu.
\newblock Adaptive {RNN} tree for large-scale human action recognition.
\newblock In {\em ICCV}, pages 1444--1452, 2017.

\bibitem[\protect\citeauthoryear{Liu \bgroup \em et al.\egroup
  }{2016}]{liu2016eccv}
Jun Liu, Amir Shahroudy, Dong Xu, and Gang Wang.
\newblock Spatio-temporal {LSTM} with trust gates for 3{D} human action
  recognition.
\newblock In {\em ECCV}, pages 816--833, 2016.

\bibitem[\protect\citeauthoryear{Liu \bgroup \em et al.\egroup
  }{2017a}]{liu2017cvpr}
Jun Liu, Gang Wang, Ping Hu, Ling-Yu Duan, and Alex~C. Kot.
\newblock Global context-aware attention {LSTM} network for 3{D} action
  recognition.
\newblock In {\em CVPR}, pages 1647--1656, 2017.

\bibitem[\protect\citeauthoryear{Liu \bgroup \em et al.\egroup
  }{2017b}]{chen2017pr}
Mengyuan Liu, Hong Liu, and Chen Chen.
\newblock Enhanced skeleton visualization for view invariant human action
  recognition.
\newblock {\em PR}, 68:346--362, 2017.

\bibitem[\protect\citeauthoryear{M\"uller \bgroup \em et al.\egroup
  }{2005}]{muller2005hdm05}
M.~M\"uller, T.~R\"oder, and M.~Clausen.
\newblock Efficient content-based retrieval of motion capture data.
\newblock {\em ACM Transactions on Graphic}, 24(3):677--685, 2005.

\bibitem[\protect\citeauthoryear{Presti and La~Cascia}{2016}]{presti20163d}
Liliana~Lo Presti and Marco La~Cascia.
\newblock 3{D} skeleton-based human action classification.
\newblock {\em PR}, 53:130--147, 2016.

\bibitem[\protect\citeauthoryear{Raptis \bgroup \em et al.\egroup
  }{2011}]{raptis2011real-time}
Michalis Raptis, Darko Kirovski, and Hugues Hoppe.
\newblock Real-time classification of dance gestures from skeleton animation.
\newblock In {\em Proceedings of the 2011 ACM SIGGRAPH/Eurographics symposium
  on computer animation}, pages 147--156. ACM, 2011.

\bibitem[\protect\citeauthoryear{Shahroudy \bgroup \em et al.\egroup
  }{2016}]{sharhroudy2016cvpr}
Amir Shahroudy, Jun Liu, Tian-Tsong Ng, and Gang Wang.
\newblock {NTU RGB+D}: A large scale dataset for 3{D} human activity analysis.
\newblock In {\em CVPR}, pages 1010--1019, 2016.

\bibitem[\protect\citeauthoryear{Song \bgroup \em et al.\egroup
  }{2017}]{song2017aaai}
Sijie Song, Cuiling Lan, Junliang Xing, Wenjun Zeng, and Jiaying Liu.
\newblock An end-to-end spatio-temporal attention model for human action
  recognition from skeleton data.
\newblock In {\em AAAI}, pages 4263--4270, 2017.

\bibitem[\protect\citeauthoryear{Wang and Wang}{2017}]{wang2017cvprtwostream}
Hongsong Wang and Liang Wang.
\newblock Modeling temporal dynamics and spatial configuration of actions using
  two-stream recurrent neural networks.
\newblock In {\em CVPR}, 2017.

\bibitem[\protect\citeauthoryear{Xia \bgroup \em et al.\egroup
  }{2012}]{xia2012cvprw}
Lu~Xia, Chiachih Chen, and Jake~K Aggarwal.
\newblock View invariant human action recognition using histograms of 3{D}
  joints.
\newblock In {\em CVPRW}, pages 20--27, 2012.

\bibitem[\protect\citeauthoryear{Zagoruyko and
  Komodakis}{2016}]{zagoruyko2016bmvc}
Sergey Zagoruyko and Nikos Komodakis.
\newblock Wide residual networks.
\newblock In {\em BMVC}, 2016.

\bibitem[\protect\citeauthoryear{Zhang \bgroup \em et al.\egroup
  }{2016}]{wang2016cvprrealtime}
Bowen Zhang, Limin Wang, Zhe Wang, Yu~Qiao, and HanLi Wang.
\newblock Real-time action recognition with enhanced motion vector {CNN}s.
\newblock In {\em CVPR}, pages 2718--2726, 2016.

\bibitem[\protect\citeauthoryear{Zhang \bgroup \em et al.\egroup
  }{2017}]{zhang2017iccv}
Pengfei Zhang, Cuiling Lan, Junliang Xing, Wenjun Zeng, Jianru Xue, and Nanning
  Zheng.
\newblock View adaptive recurrent neural networks for high performance human
  action recognition from skeleton data.
\newblock In {\em ICCV}, 2017.

\bibitem[\protect\citeauthoryear{Zhu \bgroup \em et al.\egroup
  }{2016}]{zhu2016aaai}
Wentao Zhu, Cuiling Lan, Junliang Xing, Wenjun Zeng, Yanghao Li, Li~Shen, and
  Xiaohui Xie.
\newblock Co-occurrence feature learning for skeleton based action recognition
  using regularized deep {LSTM} networks.
\newblock In {\em AAAI}, pages 3697--3704, 2016.

\end{thebibliography}

\end{document}